\def\year{2022}\relax
\documentclass[letterpaper]{article} 
\usepackage{aaai22}  
\usepackage{times}  
\usepackage{helvet}  
\usepackage{courier}  
\usepackage[hyphens]{url}  
\usepackage{hyperref}
\usepackage{xcolor}
\usepackage[title,titletoc]{appendix}
\usepackage{graphicx} 
\usepackage{natbib}  
\usepackage{caption} 
\DeclareCaptionStyle{ruled}{labelfont=normalfont,labelsep=colon,strut=off} 
\usepackage{amsmath,amsthm,amssymb}
\usepackage{xcolor}
\usepackage{balance} 

\usepackage{algorithm}
\usepackage{algorithmicx}
\usepackage{algpseudocode}
\usepackage{graphicx}
\usepackage{subcaption}
\usepackage{textcomp}
\usepackage{xcolor}

\usepackage{newfloat}
\usepackage{listings}
\floatstyle{ruled}
\newfloat{listing}{tb}{lst}{}
\floatname{listing}{Listing}

\title{Minimizing Robot Navigation-Graph \\ For Position-Based Predictability
By Humans}

\author{    Sriram Gopalakrishnan, Subbarao Kambhampati}
\affiliations{
    School Of Computing \& AI, \\
    Arizona State University

    sgopal28@asu.edu

}

\begin{document}

\maketitle
\renewcommand{\thefootnote}{\fnsymbol{footnote}}
\begin{abstract}

In situations where humans and robots are moving in the same space whilst performing their own tasks, predictable paths taken by mobile robots can not only make the environment feel safer, but  humans can also help with the navigation in the space by avoiding path conflicts or not blocking the way. So predictable paths become vital. The cognitive effort for the human to predict the robot's path becomes untenable as the number of robots increases. As the number of humans increase, it also makes it harder for the robots to move while considering the motion of multiple humans. Additionally, if new people are entering the space --like in restaurants, banks, and hospitals-- they would have less familiarity with the trajectories typically taken by the robots; this further increases the needs for predictable robot motion along paths.

With this in mind, we propose to minimize the navigation-graph of the robot for position-based predictability, which is predictability from just the current position of the robot. This is important since the human cannot be expected to keep track of the goals and prior actions of the robot in addition to doing their own tasks. In this paper, we define measures for position-based predictability, then present and evaluate a hill-climbing algorithm to minimize the navigation-graph (directed graph) of robot motion. This is followed by the results of our human-subject experiments which support our proposed methodology.

\end{abstract}


\section{INTRODUCTION}


To motivate this work, let's consider multiple humans and robots moving in settings like hospitals or restaurants. When humans and robots are moving in the same space, it can get very challenging for both to move seamlessly due to the difficulty in mental modeling the other. Yet predictability is very important for both humans and robots to navigate in the same shared space. 

This work is built on a simple premise. Given only the current position of the robot, if the robot can take many possible future trajectories, it is harder to bound the robot's future positions. This consequently makes it harder for the human to navigate in the space without potential path conflicts. Bounding the motion is possible by limiting the possible trajectories a robot can take from any position, and minimizing the areas in which the robot can move. One might ask, why not make the robot handle all the navigation effort? This is the direction of some of the existing research which has shortcomings such as the robot freezing  ~\cite{unfreeze_assitiveHumans_HRI_navig} or moving haphazardly from being unable to compute a path that conforms with the motion of multiple people moving in the space.

\begin{figure}[ht]
\centering
     \includegraphics[width=\columnwidth]{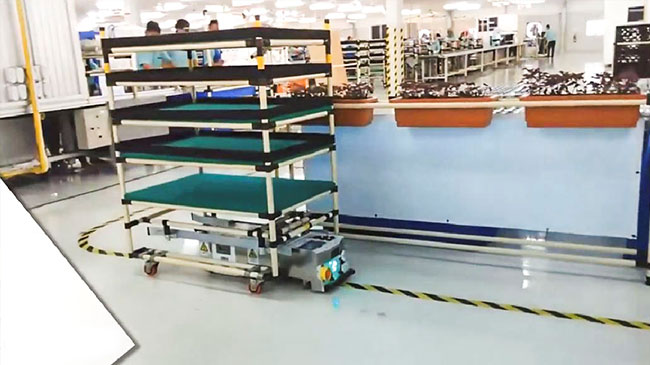}
      \caption{An Autonomous Guided Vehicle (AGV) following a path laid out in tape }
      \label{fig:agv_pic}
\end{figure}

Existing research for robot navigation in social/everyday-settings with multiple humans and robots involves a complex interplay of sensing, high level path planning, predicting human motion, and reactive behavior (eg: collision avoidance)~\cite{HRI_navigation_survey}. Navigation in larger groups of people is much harder as things like crowd dynamics need to be considered as well (~\cite{unfreeze_assitiveHumans_HRI_navig},~\cite{crowd_vasquez2014inverse},~\cite{crowd_cnn_hen2020robot}). In this work we take a different approach. If we assume the robots have a fixed set of tasks --such as serving food between the kitchen and tables, or transporting medicine between a cabinet and patient beds-- then we could compute a restricted \emph{ navigation-graph} to make navigation easier for both humans and robots in the space, and keep each task's motion path costs within a multiplicative bound of optimal path costs. Such an approach can also reduce the hardware and computation requirements for the robots in the space. Limiting motion like this is what was done with Automated/Autonomous Guided Vehicles (AGVs) which are already in use for industrial settings, and even hospitals with mature technology \cite{AGV_book_fazlollahtabar2015autonomous}. AGVs followed a predefined path, often laid out with tape (like rails for a train) to move between positions (see Figure \ref{fig:agv_pic} source \cite{CASUN_AGV}). The AGV's grid (which determine what paths can be taken) were laid out for the sake of optimizing task output. In this paper, we propose computing the AGV grid (as a directed graph) for more predictability of trajectories by humans. Predictability is very important for human robot interactions \cite{HRI_navigation_survey} and more predictability would improve adoption of AGV's in everyday settings like restaurants. 


\begin{figure}[ht]
\centering
     \includegraphics[width=\columnwidth]{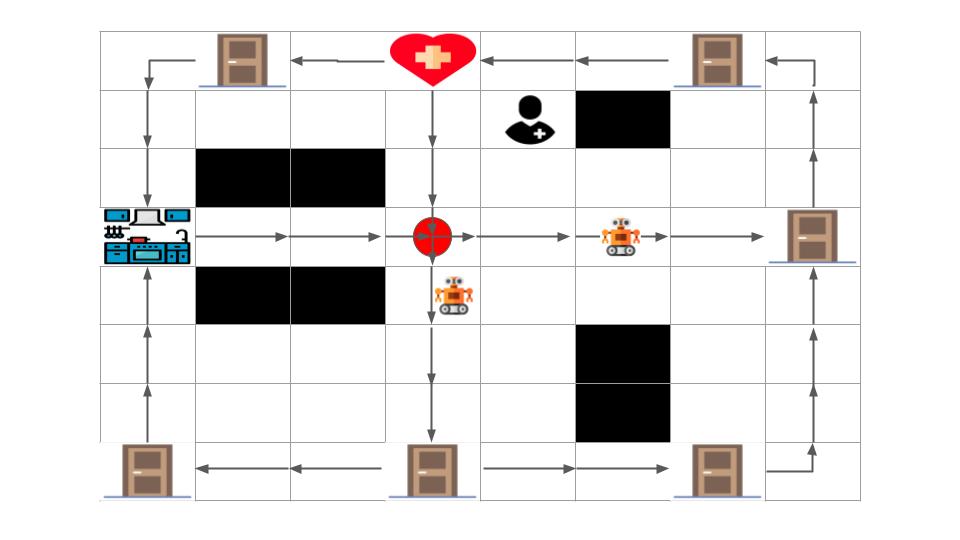}
      \caption{A hospital floor with robot navigation-graph that connects the kitchen and medicine cabinet to all patient rooms. The grid squares are the vertices of the navigation-graph, and the directed lines are the edges; only one branching-vertex (in red) has more than one outgoing edge}
      \label{fig:nursing_home_predictable}
\end{figure}
\begin{figure}[ht]
\centering
     \includegraphics[width=\columnwidth]{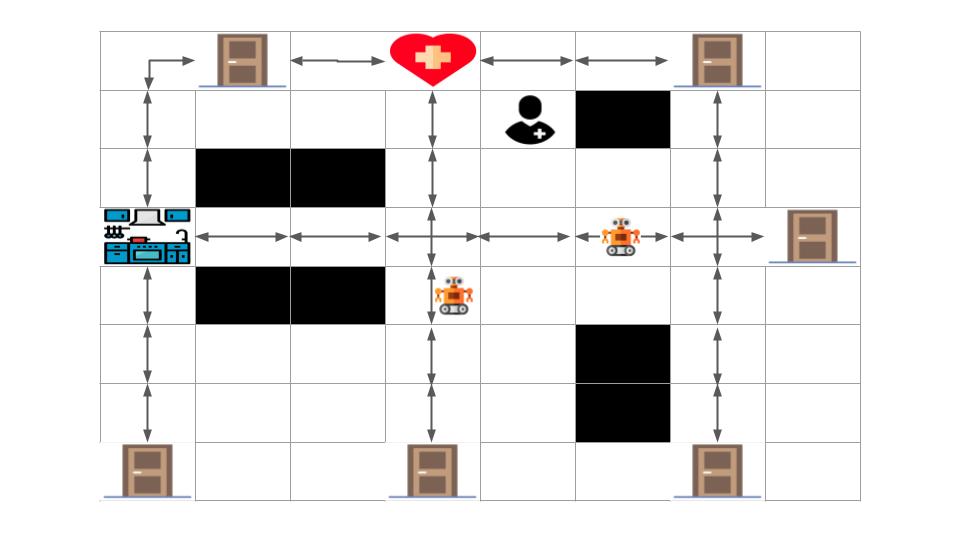}
      \caption{A hospital floor with robot navigation-graph that uses only the shortest paths to go between kitchen and medicine cabinet to all patient rooms. The grid squares are the vertices of the navigation-graph, and the lines are the edges; every vertex/position is a branching-vertex in this example.}
      \label{fig:nursing_home_shortest_paths}
\end{figure}

Constraining the robot's navigation-graph --where and how robots can move (like in Figure \ref{fig:nursing_home_predictable})-- and communicating it to the humans in the space by markers or tape can help in two ways. One way is by simplifying the path planning of the robot; it is now helpfully limited to computing paths on the navigation-graph. Another way it helps is by making the robot's motion more predictable to the humans. It does this by limiting the space the robots can occupy and how they can move. This helps reduce or negate the need for the human to mental model the robots as they know the robot is limited to following the laid out markers. Humans that know the path of the robots tend to help in the navigation by simply not blocking or adapting their movement to avoid conflicts, as humans in the space can be seen as non-competitive or collaborative. This idea of modeling the human as working with the robot in navigating the space was used in ~\cite{unfreeze_assitiveHumans_HRI_navig},~\cite{coophuman_HRI_navig},~\cite{mobilerobots_populatedEnv}, ~\cite{assitiveHumans_HRI_navig2}. The involvement of the humans in effective robot navigation is almost necessary when dealing with many humans and robots in the same space as the robot might freeze ~\cite{unfreeze_assitiveHumans_HRI_navig}~\cite{trautman2015robot_crowd} or move haphazardly because it was unable to compute a path that is conformant with the motion of all humans in the space. This is why predictability of robot path is very important, as it enables easy and effective involvement of the humans in the space. This work focuses on the computation of the navigation-graph for robots to make predictability as easy as possible, and with minimal information. Minimal information and effort is important as we cannot expect the human --who has their own tasks-- to invest non-trivial cognitive effort in predicting motion. This is why we focus on \emph{position-based predictability}, where the future trajectory of the robots can be predicted or bounded from the current position alone. Since predictability is deemed to be a very desirable quality in robot navigation from a human factors perspective \cite{HRI_navigation_survey} it would affect adoption or acceptance of robots in any setting.



In Figures \ref{fig:nursing_home_shortest_paths} and \ref{fig:nursing_home_predictable} a nurse is moving in the space with robots. Certain regions are blocked out (black) due to furniture, walls or other motion constraints. If the nurse (in the example) is moving down towards a patient's room in the middle, then in the case shown in Figure \ref{fig:nursing_home_shortest_paths} the nurse wouldn't know how the robots would move and cannot easily come up with a non-conflicting path. The robots might also reverse their direction if their task priorities change. In Figure \ref{fig:nursing_home_predictable} the nurse can easily infer that the robots will only be moving away and can walk straight to the patient door. This is because the navigation-graph has only 1 ``Branching-Vertex" (position), which are vertices with more than one outgoing edge. In Figure \ref{fig:nursing_home_shortest_paths}, every position is a branching-vertex for the robot's navigation. Fewer branching vertices mean fewer possible trajectories from any position and so makes prediction or bounding of robot motion easier. This comes at the expense of path costs between positions associated with tasks (called terminal vertices). Reducing branching vertices while keeping path costs within acceptable bounds are the main objectives of our approach.

In this paper, we formalize the problem of minimizing the navigation-graph for position-based predictability. We then introduce measures for position-based predictability, followed by a hill-climbing algorithm to minimize the graph and optimize for position-based predictability. We then discuss the experiments that we did to show the validity of our approach, followed by human subject experiments to check our intuition that minimizing branching vertices helps predictability.

\section{RELATED WORK}

In the area of robot motion planning with humans in mind ~\cite{ancaRss}~\cite{ancaThesis}, Dragan et al. assume that the robot has a model that can help predict what motion the human will infer, and move predictably according to that model. Along a related line of thinking \cite{knight2016expressive} considers how the robot's actions communicate intent to the human, and use it for better collaboration. An implicit assumption in such work is that the human pays non-trivial attention to the actions of the robot to make this inference, which is defensible when the human and robot are doing the same or related tasks, or in a sparse setting like at home. In our settings of robot motion (hospitals, restaurants, banks), we expect the humans will at most glance at the robot or use peripheral vision to determine where it is (not invest time intently observing the robot); the humans would be preoccupied with their own independent goal-directed behavior. There is also work \cite{broz2013planning} that considers inferring the hidden mental state of humans to infer their intent with POMDPs and use that to coordinate better. In all of the aforementioned work, if there are many people moving in the space, then it may be computationally intractable or just not possible for the robot to act and conform to all of the humans' intents or mental models. This leads us to the work on robot motion around larger groups of people.

In the literature that tackles robot navigation in groups/crowds, the approaches taken include modelling group dynamics using Gaussian processes \cite{unfreeze_assitiveHumans_HRI_navig} to infer how the crowds will move use it for navigation. Naturally this would require more hardware to sense and compute crowd dynamics and it is not clear how effective this actually is; to our knowledge, the approach has only been tested on constrained simulations. A more pressing concern is if there are any useful or predictable crowd dynamics in settings like hospitals. Such dynamics may exist in busy pedestrian crossings where there are general directions of motion. We think such predictable dynamics are unlikely to exist in settings like hospital where movement is dictated by individual hospital staff goals which can be very hard to predict. Another concern is how seamless the motion will be as the predictability of motion is known to be important from a human factors perspective ~\cite{HRI_navigation_survey}, this was not evaluated in the referenced work on crowd modelling with Gaussian processes \cite{unfreeze_assitiveHumans_HRI_navig}. We expect people moving around the robot without consistent dynamics would result in jerky, inconsistent movement. This leads to inconsistent signals given to the human, and hurt navigation coordination with the human. If robot goals or priorities can change dynamically, then the predictability of motion is worsened. In the approach we support, by restricting the robot to simply following the markers on the floor, which communicates a directed navigation-graph, there is much less uncertainty on how the robot will move. The only source of uncertainty is in how the robots will move/turn at branching vertices, and by minimizing the number of these, we further improve predictability.

Other methods for navigation in crowds include using Inverse reinforcement learning ~\cite{IRL_crowd_navig}~\cite{IRL_vasquez2014inverse}, and Deep reinforcement learning techniques ~\cite{deep_rl_chen2020robot},~\cite{deepl_rl_attention_chen2019crowd} which also consider human gaze in the feature set. With IRL approaches, if motion dynamics of people change --which can happen with changes in furniture and space rearrangement-- then the relearning time is not considered in such work. With our method, if the space configuration changes do not overlap with the navigation-graph, we can safely ignore it. If they do, we simply update the initial graph with the new space configuration (and capture which vertices and edges are blocked), and compute a new navigation-graph and replace the old one's markers/tape. Additionally, it is not clear how predictable the motion with IRL techniques is, as that aspect was not evaluated. Similar problems like motion predictability and robustness to environment reconfiguration would plague deep-RL methods(~\cite{deep_rl_chen2020robot},~\cite{deepl_rl_attention_chen2019crowd}). In these, experiments were done in simulations, and no consideration was given to the predictability of the motion by humans. In contrast, AGV navigation using tape laid on the floor has a history of real-world use in hospitals and industrial settings. To bring this approach to more settings, we take the position that further improving predictability of motion is paramount for adoption in more commercial settings. Our work is about how to compute the navigation-graph that is laid out, such that motion is very easily predictable. This allows the humans to move easily in the space even with multiple robots, as they know the robot's hard motion constraints; they do not have to build much trust in the robot's navigation and collision avoidance abilities. This is likely to matter more with bigger robots, as they are perceived as more of a threat to physical safety ~\cite{bigRobots_hiroi2008bigger}. On top of improving predictability, one can incorporate any additional advancements for soft robotics ~\cite{softRobots_arnold2017tactile}, collision avoidance, and human gaze detection for use in further improving the safety and acceptability in everyday settings.

Our approach also connects to recent literature on environment design, in which the environment is configured to make prediction by humans easier ~\cite{designForInterp}. Obviously, we do not explicitly reconfigure the environment; we only limit the environment in the robot's model for its motion. In environment design work on predictability with respect to the human, one assumes the goal is known to the human, or that the human has observed a prefix of a plan's actions and can infer the rest assuming they can infer optimal path completions. These are assumptions that we want to remove for the settings we consider (restaurants, banks, hospitals). Humans in such spaces are not paying significant attention to the actions or know the set of goals of the (possibly) multiple robots moving, nor can they be expected to invest effort computing optimal path completions to these goals. There can also be new humans entering the space regularly with no familiarity. So making predictability as trivial as possible, and from just the current position (as AGVs do) becomes important. 

With respect to the computation and algorithmic aspect of our work, our work computes a navigation-graph in a problem setting similar to Strongly Connected Steiner Subgraph(SCSS) ~\cite{SCSS_problem} problem. In SCSS, given a directed graph and a set of Terminal nodes, the objective is to find a minimally weighted subgraph (can consider edge weights and/or vertex weights) such that the terminal nodes are strongly connected, i.e. there is a path between each pair in both directions. In comparison, our optimization for position-based predictability is done with two objectives: (1) minimize the vertices and edges (i.e. space) used by the robot in it's navigation between terminal vertices; (2) minimize the number of branching vertices, and thus possible paths that a robot can take from any position. The former is what the SCSS problem does, but to the best of our knowledge no variants of SCSS explicitly considers minizing branching vertices. There are graph problems that consider branching vertices, but with only a single source node (as far as we know). Specifically, the problem of Directed Steiner Tree with Limited Diffusing vertices (DSTLD), which has applications in multicast packet broadcasting ~\cite{degrConstrDirSteinerTree}. Solutions to DSTLD optimizes for the total edge-weight cost of the tree (Steiner tree problem), \emph{and} limits the number of branching vertices (which they call diffusing vertices) but only from a single source node. They do not consider the problem of optimizing the number of branching vertices for paths between multiple source and destination vertices (which we do). There is also no thought given for individual path costs between vertices (only total edge weight). Another such problem in this vein of literature, is the Rectilinear Steiner Arborescence (RSA)~\cite{RSMA_rao1992rectilinear} which is the rectilinear version (manhattan distances) of Directed Steiner Tree problem . In RSA, the objective is to compute the minimum length (sum of edge cost) directed tree, that is rooted at an origin point and connected to N vertices. The problem is known to be NP-complete ~\cite{RSA_NPcomplete_shi2000rectilinear} to compute a solution that is less than a specified length, and so NP-hard to optimize (minimize) for the total length. RSA (like DSTLD) is just for a single-source, and pays no heed to pairwise path costs between terminals.

\section{PROBLEM FORMULATION}
The problem of graph minimization for position-based predictability is given by the tuple
\begin{equation}
P_{min} = \langle G,T,C,W\rangle
\end{equation}

where
\begin{itemize}
    \item \emph{G} is the directed graph defined by its vertices and edges \emph{(V,E)}. This graph captures the all the allowed motion of the robot. Any motion constraints like protected areas or one-way corridors are captured in the graph as well (dropped vertices, and one-way edges).
    \item \emph{T} is the set of ordered vertex pairs that need to remain connected in the minimized graph. Each pair appears twice (both orderings of the pair) to represent both directions of connectivity. The vertices will be called terminal vertices, and we refer to the ordered pair of vertices as a ``task", as we think of them as being associated to a task. These are locations of importance for a task like a medicine cabinet, or it could be the door connecting to an adjacent room that the robot needs to reach.
    \item \emph{C} is the function that returns the cutoff cost (maximum allowed) distance associated to each task in \emph{T}. The cost of the paths in the final minimized graph should be less than the specified corresponding cutoff.
    \item \emph{W} is a function that returns the weight of each pair in \emph{T}. This weight can be the probability of the task, or an importance weight (if some tasks need to be done more quickly). 
\end{itemize}

The objective is to reduce the input graph \emph{G} so as to improve position-based predictability, which in this work we take to be predictability from \emph{just} the current position of the robot. Measures for this position-based predictability are formalized in the following section. Note that we do \emph{not} expect the human to predict the entire path suffix from the current position of the robot; people certainly do not predict other pedestrians' full trajectories before moving. Rather, we want to make easier the bounding of possible paths or positions that the robot might occupy. This information would help the human decide where to move for their next steps, and if the robot is likely to move in the same spaces. 

\subsection{Measures for Position-Based Predictability}
We describe herein measures for position-based predictability and justify their definitions. 

\textbf{ Weighted Prediction Cost (\emph{WPC}):} If the navigation-graph as fewer branching vertices, and each of them having a smaller branching factor (number of out-edges), then it makes the motion more predictable for the human; fewer possible future robot trajectories for the human to consider. Vertices with only one outgoing edge require minimal thought, and trajectories with a sequence of vertices with just one outgoing edge are trivial to predict from just the current position of the robot; these are ideal conditions for predictability (trivial cost). 

We first describe the WPC measure before defining it. There are two components in WPC, the first counts the number of branching vertices that appear on the paths for the robots tasks, and weights it according to the values given in $W$. This is then multiplied by the sum of the branching factor of those branching vertices. Two terms are necessary as graphs might have the same sum of branching factors but different number of branching vertices or vice versa. We need both the number of branching vertices to be fewer \emph{and} their branching factor to be smaller to make predictability easier. 3 branching vertices with branching factor 2 is treated as worse than 2 branching vertices with branching factor 3. This is because every branching-vertex breaks the ideal of just one trajectory for as many steps as possible. If the human has to cross one of the branching paths that the robot is might take, the human may slow down to see how the robot turns(assuming they want to avoid conflict like in ~\cite{unfreeze_assitiveHumans_HRI_navig},~\cite{coophuman_HRI_navig}). Such a situation would occur more often with more branching vertices. So the count matters as well as the sum of branching factors (number of trajectories). The branching factor matters because fewer trajectories lets the human plan a path around any of the future trajectories more easily (less space covered). A smaller WPC implies a simpler graph of robot motion for position-based prediction. Zero WPC, meaning zero branching vertices implies the robots movements are all trivially predictable since there is only one action (direction) it can take from every position. WPC is defined as :
\begin{multline}
    WPC(G,T,W) =
        \sum_{t \in T}\sum_{v \in SPV(G,t)}W(t)*1[deg^+(G,v)>1 ]  \\
        \times \sum_{t \in T}\sum_{v \in SPV(G,t)}W(t)*deg^+(G,v)
    \label{eqn:WPC}
\end{multline}

where $SPV(G,t)$ returns the shortest path vertices for the input task (t) in the graph (G) being evaluated, and $deg^+(G,v)$ returns the outdegree of the vertex(v) in the graph(G). If there are multiple shortest paths for a task, then the path that contributes the least cost to WPC is used. What this translates to--in terms of the robot's motion--is that when the robot is moving on the path associated to $T$, the human observing it at any position will have to consider fewer (ideally only 1) possible future trajectories when trying to move and avoid any conflicts. 



\textbf{Weighted Ratio \emph{NV/NBV}:} where NV is the number of vertices and NBV is the number of branching vertices. This measure represents the number of vertices or steps in a sequence with no branching, and is weighted by the task weights. If the weights are just the probabilities of the tasks, then this measure returns the average number of steps before encountering a branching-vertex. So a larger NV/NBV value implies longer paths with no branching, i.e. going one-way for longer distances; this makes path predictions using only the position trivial for many parts of the graph. NV/NBV is formalized as:

\begin{multline}
    NV/NBV(G,T,W) = \\
    \frac{\sum_{t \in T}\sum_{v \in  SPV(G,t)}W(t)}{\sum_{t \in T}\sum_{v \in SPV(G,t)}W(t)*1[deg^+(G,v)>1 ]}
\end{multline}

What this means for the human in the space, is that they don't need to consider multiple paths for moving shorter distances. They just need to see that in the next "N" steps or units of time, the robot will not be near the human. This was illustrated in Figure \ref{fig:nursing_home_predictable}. If the nurse was moving to the middle door at the bottom, then the human knows with a glance that the robots will only ever move further away from the human and never towards them, or cross them (as per the navigation-graph). The human doesn't need to infer the full path of each of the robots, which would require unnecessary effort. Even knowing both the robots' goals doesn't necessarily make the cognitive effort minimal. For the human (the nurse) to reach their goal (middle door) they just need to know or bound the immediate next few steps to determine a uninterrupted path to their goal (middle door). Any additional effort is unnecessary, and thus undesired.

\begin{figure}[ht]
\centering
     \includegraphics[width=\columnwidth]{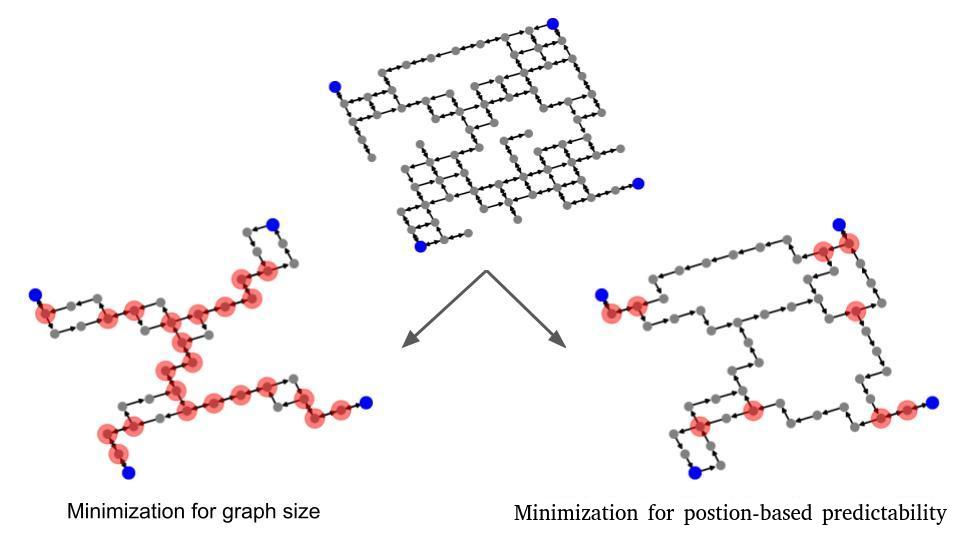}
      \caption{Example of graph minimization for graph size, and minimization for position-based predictability. Blue vertices are the terminal vertices that must stay connected, and vertices highlighted in red are branching vertices}
      \label{fig:example_minimiz}
\end{figure}

An example of the type of navigation-graph optimization that this work does is shown in the rectilinear grid-graph in Figure \ref{fig:example_minimiz}. Thee blue vertices are the terminal vertices and vertices highlighted in red are branching vertices. In the figure, we show two graph minimizations; one that focuses only on graph size (number of vertices and edges), and another minimization that optimizes for position-based predictability (fewer branching vertices and graph size).

Both these measures involve branching vertices. We provide supporting evidence for our intuition in the human subjects experiments that fewer branching vertices make position-based predictability easier. 


\section{METHODOLOGY}


Our problem of (navigation) graph minimization for position-based predictability is one of constrained optimization. We want to find a graph that minimizes WPC and has a high value for NV/NBV, while connecting all the required pairs of terminal vertices in \emph{T} within the cost constraints specified through \emph{C}.
To find such a minimized graph, we used a hill-climbing search approach. The two key components in our approach are (1) finding a diverse set of paths for each task (ordered pair of terminal vertices), and (2) evaluating (for comparison) the different graph candidates during the hill climbing search. The pseudocode is in the supplemental text.

Before we go into the details, we highlight a challenging aspect of this problem which motivates our methodology choice. Specifically, the number of possible combination of paths between terminal vertices for a graph can explode in size based on the number of vertices and graph-connectivity (denser = more path options). The difficulty in optimization comes from the fact that we cannot choose a path between terminal vertices independent of the other paths. This is because the number of branching vertices is not known until we combine the paths for all pairs of terminal vertices. This leads to a combinatorial explosion in the number of possible path combinations (and thus graphs), and so makes it challenging to compute the optimal graph. 

To make the computation tractable, we limit ourselves to using a fixed number of paths for each task in $T$. This initial population of paths determines the quality of the outcome of the search, and choosing this population is the first step of our approach. In this regard, one set of helpful constraints are the cost cutoffs for the paths. This helps constrain our search space in terms of the paths to consider. The paths are obtained through shortest-path search for each ordered pair of vertices in $T$. After one path is found for a task, the weights of those edges are doubled and the process is repeated to help build a diverse set of alternative paths. After we build the candidate pool of paths per task, we start the hill-climbing search.

This brings us to the other part of the search process, and that is generating and comparing successive graph candidates. For generating new candidates for a given graph in the search, we only consider single-path (associated to a single task) replacements to the graph. We do this using the population of paths we stored in the first step. This is how new candidate graphs are generated during the search. We may replace or just drop a path if there exists another path in the candidate graph --using edges from other paths-- that is within the cutoff. If a replacement breaks the connectivity between any terminal vertex pair, it is not considered.We compare the cost of all the candidate graphs produced at each step, and greedily take the candidate with the lowest cost (we will discuss evaluation functions used for comparison soon). If no better candidate exists at a step in the search, we stop the search process. Lastly, we use a fixed number of random restarts (5 restarts) to repeat the hill climbing process, and use the best result over all. Each iteration starts with a graph that is a combination of randomly chosen paths for each task. The described algorithm's pseudo-code is given in the supplemental material.

As maybe apparent, the cost or evaluation function is pivotal to the algorithm. We test our hill climbing approach with two relevant cost functions that are intended to minimize for the effort of an observer for position-based predictability. We start with discussing the baseline cost function, which we then modify for our proposed (main) cost function. The baseline cost function is simply the sum of weighted vertices and edges.  We will call this the \emph{Graph-size cost} or GSC, so called since a smaller graph with fewer vertices and edges would have a lower cost. GSC is analogous to the optimization objective of the SCSS problem (see related work). This was part of the motivation for setting GSC as the baseline, as well as the fact that making the navigation graph as small as possible ought to help predictability; intuitively a smaller graph implies fewer possible paths. We could not directly compare with any algorithms for SCSS from the literature as the papers associated to the problems were only theoretical papers and no code was written; we reached out to the authors and confirmed this. 

One difference between how we use GSC and the SCSS problem is that SCSS does not consider individual path costs between terminal nodes, which is important for our navigation problem; during search in graph space we ensure that the path between any pair of terminal nodes doesn't exceed the specified cutoff. The other difference with SCSS, is that the weights of the nodes and edges is affected by the weight of the tasks on whose paths their appear in (in the computed subgraph). So the weights are dynamic unlike in SCSS. The importance of this is apparent when we consider the task weights as task frequency. The paths associated to more frequent tasks are taken more often, and so those paths should be more predictable. The weights are given in the problem input $W$.


\begin{multline}
    GSC(G,T,W) =
    \sum_{e \in E(G)}\max_{\{t|t \in T, e \in SPE(G,t)\}}W(t) +\\
    \sum_{v \in V(G)} \max_{\{t|t \in T, v \in SPV(G,t)\}}W(t)
\end{multline}
where $V(G)$ returns the vertices in the graph, $E(G)$ returns the edges in the graph, and $SPE(G,t)$ returns the shortest path edges for the task(t) in the graph(G), and  $SPV(G,t)$ returns the shortest path vertices for the input task (t) in the graph (G).

We use the maximum of the task weights of the tasks whose shortest path uses that edge or vertex. This encourages the search to prefer graphs that reuse the same vertex or edge in the paths for multiple task. The drop in cost would be greater when paths of tasks with higher weights use the same edges and vertices. Note that while GSC doesn't explicitly consider branching vertices, it would implicitly favor fewer branching vertices when possible; a branching-vertex increases the number of edges for every step after it, as opposed to if the path didn't branch. This would increase the edge cost part of the GSC score. So any approach to minimization for position-based predictability should be no worse than this baseline's performance for the predictability scores defined in the problem formulation section.This brings us to our cost function which builds on GSC to explicitly incorporate branching vertices. We will refer to this cost function as \emph{Branching-Vertices Cost} (BVC). The function is as follows:
\begin{equation}
    BVC(G,T,W) = WPC(G,T,W)*GSC(G,T,W)
\end{equation}
where WPC is as defined earlier in Equation \ref{eqn:WPC}

Graph minimization with BVC as compared to minimization with GSC will tell us if simply minimizing the graph for the number of vertices and edges (GSC) is sufficient to optimize for our position-based predictability measures or better in practice (human-subject studies). If so, then our BVC cost, explicitly considering branching vertices would not be necessary which is the premise of this work. More importantly, GSC gives us a plausible baseline to use in human subject experiments for position-based predictability. It represents the stance that minimizing just for the size of the graph is enough and ignoring branching vertices doesn't hurt predictability.


Note that we do not directly optimize to increase $NV/NBV$ as that biases the search to use longer paths in the resultant graph for getting a higher value in this measure. The cutoff input in $C$ is the \emph{upper limit}, but we think it desirable to keep the paths as short as possible. So we do not explicitly incorporate the $NV/NBV$ measure into the cost.


\subsection{Computational Complexity}

For our problem over rectilinear graphs, it is hard to compute optimal solutions as even the special case with all tasks having the same (single) source vertex is NP-Hard to optimize since that special case is the Rectilinear Steiner Arborescence (RSA) problem as mentioned in the related work. So this problem is atleast as hard as the RSA problem. 

With respect to our algorithm's complexity, for the first step of computing the population of paths for each task in $T$, the entire subroutine should take $O(PopulationSize*(|E|+|V|log|V|))$; where $(|E|+|V|log|V|)$ is from the running time of shortest path using Dijkstra's algorithm. 

The hill-climbing part of our algorithm is an anytime algorithm. We can stop at any point and get the best solution found until then. With more random restarts and letting the algorithm run longer, one can get better solutions as is expected with hill-climbing. This computational cost is infrequent, and only needed if the space configuration changes. A helpful factor is that a lot of the computation steps can be effectively parallelized, such as the steps of constructing and evaluating the next candidate state. Additionally, each iteration of hill-climbing can be done in parallel. This can help deal with very large graphs.

\section{EXPERIMENTS}

To evaluate the graph minimization with our hill climbing algorithm, we run our algorithm on randomly generated 20x20 grids. This is intended to reflect a room or hall of a building. The robots have to move through this space between terminal vertices; terminal vertices could be doors connecting to other rooms, or locations relevant to a task. We start with a fully connected grid with adjacent vertices connected bidirectionally. All edges are of unit distance. Then we randomly drop 20\% of the vertices and 20\% of the remaining edges; dropping vertices and edges represents the spaces used for furniture, obstacles, corridors, walls etc. Lastly we arbitrarily select the terminal vertices from the remaining vertices. All ordered pairs of terminal vertices define the set $T$. We vary the number of terminal vertices selected from the set $\{3,4,6,8\}$. When this parameter is \emph{not} being varied (in the subsequent plots), the default value is 6. Another parameter that we vary in our experiments is the cutoff cost $C$ for the path costs between the terminal vertices. The cutoff cost is set as a multiple of the shortest path cost for each pair of terminal vertices, so it is a bound on suboptimality. If the cutoff is $2$, then only paths that are less than or equal to twice the optimal cost are considered. In our experiment we vary this value from the set $\{1,2,3,5\}$ where 1, means only optimal paths are considered. When this parameter is \emph{not} being varied (in the subsequent plots), the default value is 3. Lastly, the weights $W$ are randomly assigned to all tasks in the range $[0,1)$ and normalized so they would sum to 1. In total, the experimental settings include every combination for the number of terminal vertices and cutoff bound. For each unique setting, we generate 10 random graphs (setting the random seed in the code from 0 to 9) and ran the algorithm with both cost functions on the same graphs. This allows us to get a range of predictability measures produced by the two approaches for the same setting.

The maximum number of candidate paths for each $t \in T$ during the search process was capped at 20; there may not exist as many paths within a specified cost cutoff, and so it is the maximum number. We will refer to this as the  \emph{Population-Size} parameter.

The algorithm was programmed in Python, and using  networkx \cite{networkx} for graph operations including generating random graphs, and shortest path computations. The experiments were run on a PC with Intel® Core™ i7-6700 CPU, running at 3.40GHz on Ubuntu 16.04 with 32 GB of memory. The code to execute our algorithm will be available along with scripts to reproduce the data, a readme file with instructions and dependencies, and the raw data for the results in this paper.

\section{RESULTS AND ANALYSIS}
\begin{figure*}
\begin{subfigure}{.5\textwidth}
  \centering
  \includegraphics[width=\columnwidth]{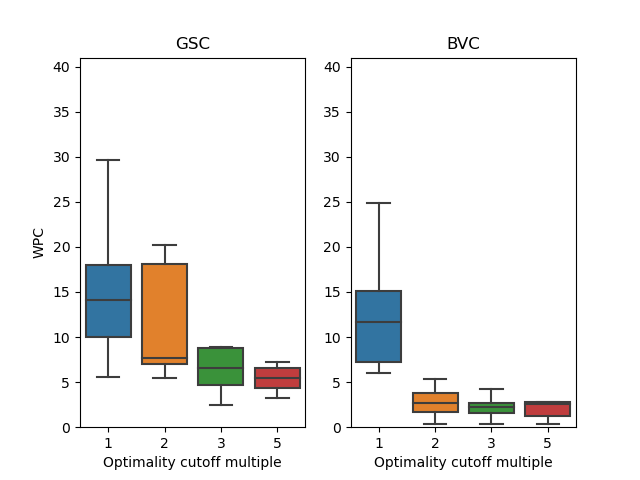}
  \caption{WPC measure with varying optimality cutoff}
  \label{fig:wpc_optimality}
\end{subfigure}%
\begin{subfigure}{.5\textwidth}
  \centering
  \includegraphics[width=\columnwidth]{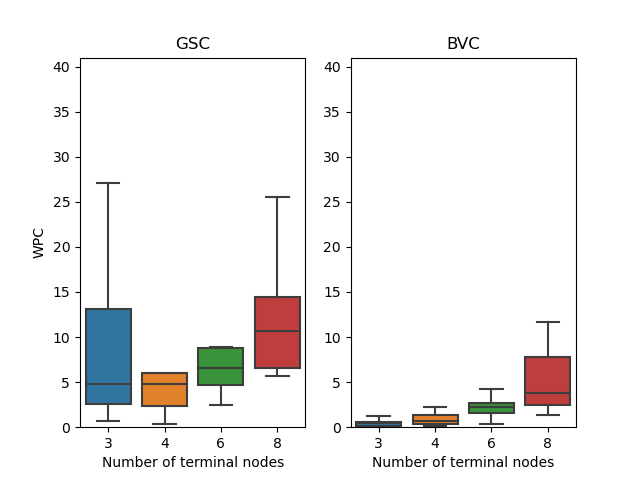}
  \caption{WPC measure with varying number of terminal vertices}
  \label{fig:wpc_terminal}
\end{subfigure}
\begin{subfigure}{.5\textwidth}
  \centering
  \includegraphics[width=\columnwidth]{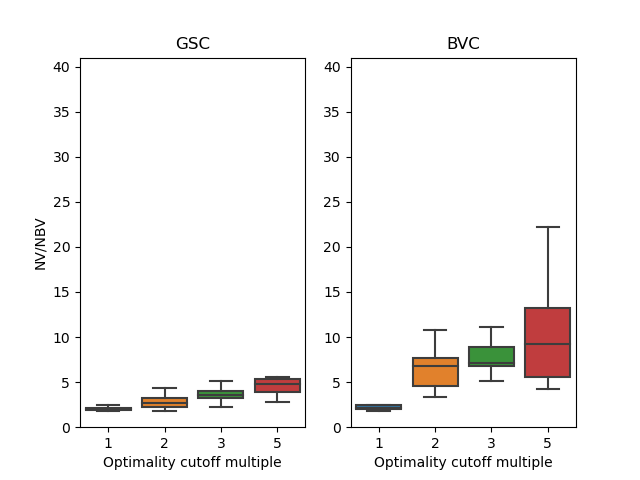}
  \caption{NV/NBV with varying optimality cutoff}
  \label{fig:NV/NBV_optimality}
\end{subfigure}%
\begin{subfigure}{.5\textwidth}
  \centering
  \includegraphics[width=\columnwidth]{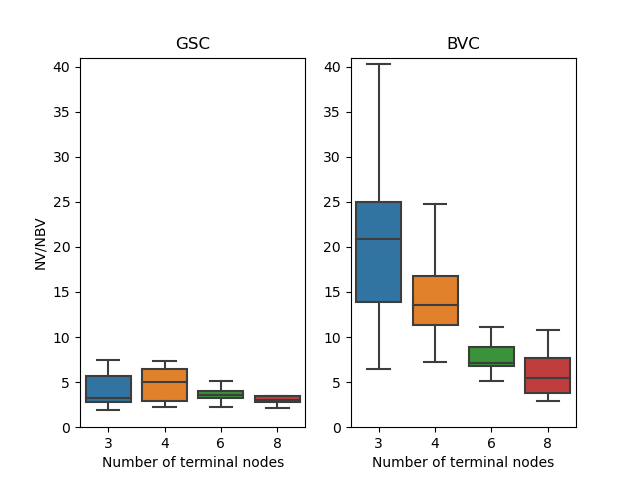}
  \caption{NV/NBV with varying terminal vertices}
  \label{fig:NV/NBV_terminal}
\end{subfigure}
\caption{Comparison of position-based predictability measures for GSC and BVC cost functions}
\label{fig:boxplots}
\end{figure*}

We present the results in Figure \ref{fig:boxplots} showing the results for the two cost functions (GSC and BVC) side by side for each measure and the same dimension of variation. Each boxplot represents the variation for that measure in that setting. Recall that for the $WPC$ measure, lower is better, and for $NV/NBV$ higher is better.

In Figure \ref{fig:wpc_optimality} and \ref{fig:NV/NBV_optimality} we vary the suboptimality of paths allowed in the search. For those plots, the number of terminal vertices is kept constant at 6. The plots show position-based predictability measures in general improve as we allow for more suboptimal paths, i.e. the $WPC$ measure decreases and $NV/NBV$ increses. This is expected as there are more possible paths for the search algorithm to work with. The BVC cost function does consistently better than GSC for all settings.

In Figure \ref{fig:wpc_terminal} and \ref{fig:NV/NBV_terminal}, we vary the number of terminal vertices while keeping the maximum suboptimality of allowed paths at 3. The position-based predictability measures worsen with more terminal vertices. This is unsurprising as the search process has to satisfy more task paths in $T$ which increases the likelihood that more branching vertices are needed since non-intersecting paths within the cutoff may not be possible. We see that the BVC cost function performs better than GSC here as well. 

GSC does improve position-based predictability measures with an increase in allowed suboptimality. Reusing vertices and edges for different paths, (which GSC encourages during the search) can reduce the number of branching vertices. However, as the data shows, it pays to explicitly factor in the branching vertices into the cost since using the BVC cost consistently gives better results in position-based predictability measures.

In Figure \ref{fig:example_minimiz}, we see one example of the typical case where GSC cost minimizing does worse than BVC for position-based predictability; BVC results
in a much more predictable navigation-graph, similar to the images of the hospital setting in Figures \ref{fig:nursing_home_shortest_paths} and \ref{fig:nursing_home_predictable}.

\begin{figure}[ht]
\centering
     \includegraphics[width=\columnwidth,]{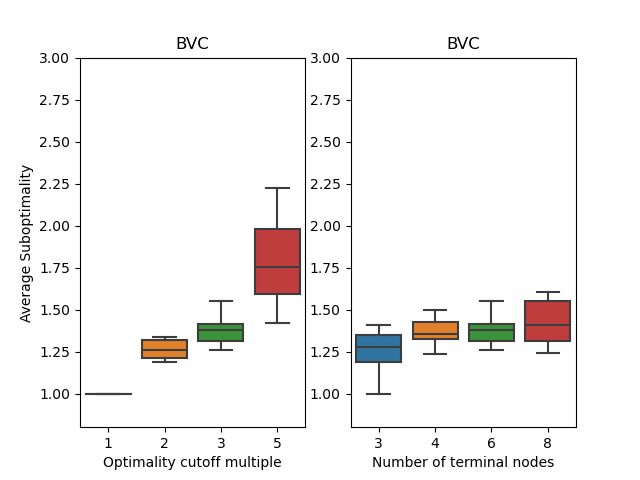}
      \caption{Average suboptimality of paths chosen with increasing allowed-suboptimality of candidate paths for GSC and BVC cost functions}
      \label{fig:avg_subopt}
\end{figure}

One might be concerned that the reason BVC does better in the $NV/NBV$ measure when the suboptimality allowed increases, is because longer paths are being abused. In Figure \ref{fig:avg_subopt}, we show the average suboptimality of the paths in the minimized graph using GSC and BVC while varying the allowed suboptimality of the initial candidate paths. The number of terminal vertices is fixed at 6. While the average suboptimality does increase, BVC keeps it lower than 2-suboptimal, and is comparable to GSC. This is because BVC (like GSC) considers the edge cost; so longer paths are avoided as they would increase the cost considerably. So the $NV/NBV$ measure is improved more by the fewer number of branching vertices, and not by the use of very long paths.

\balance
\section{HUMAN-SUBJECT EXPERIMENTS}

\begin{figure}[ht]
\centering
     \includegraphics[width=0.9\columnwidth]{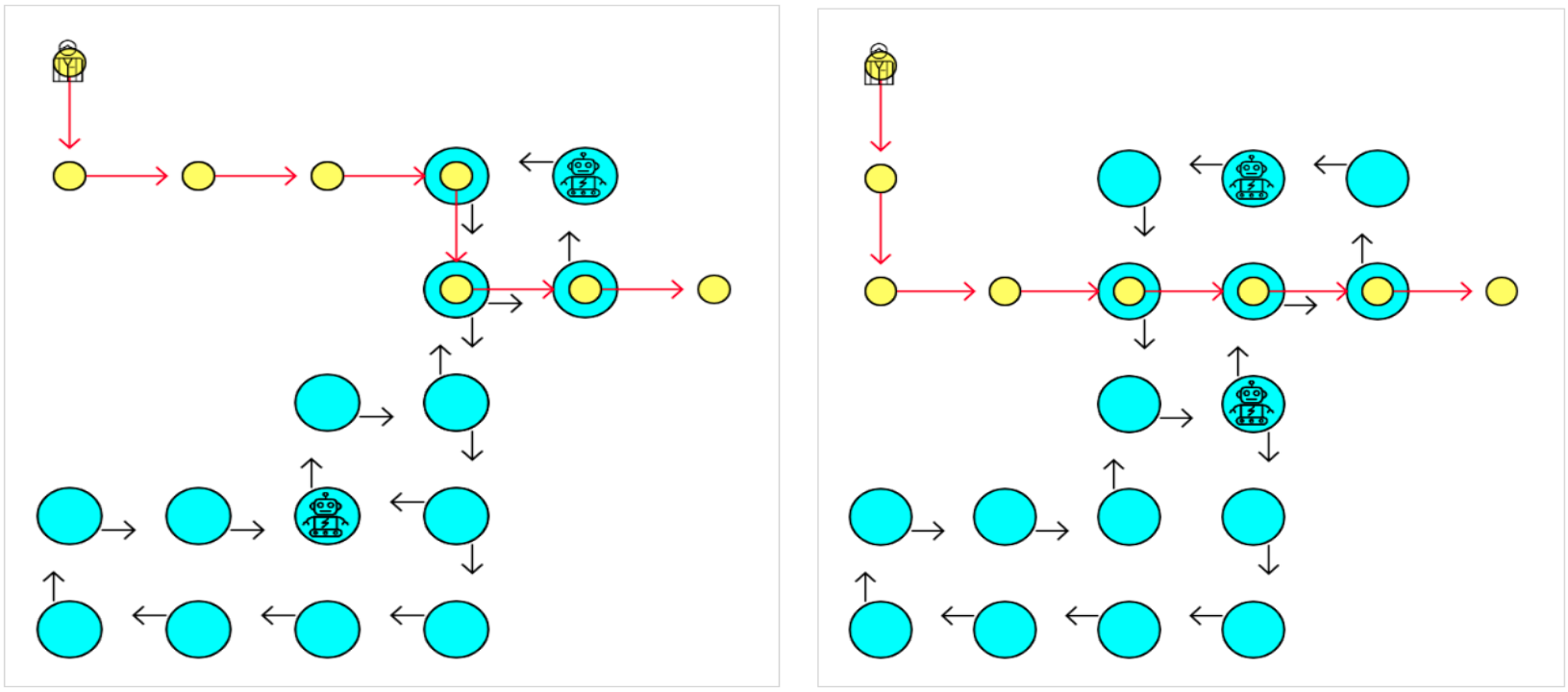}
      \caption{Problems given in the human subject experiments; problem 1 (left) has more branching vertices and problem 2 (right) has fewer.}
      \label{fig:human_studies_problem}
\end{figure}

In our human-subject experiments, we sought to evaluate if minimizing branching vertices really does make a difference in ease of predictability (from current position only). The problems we presented to the participants were of a 6x6 grid with 20\% of the edges removed (along with any orphaned vertices). There were two robots in each problem, and a path defined for the human to move, as illustrated in Figure \ref{fig:human_studies_problem}. The robots moved on the blue circles according to the black arrows, and the human moved on the yellow circles according to the red arrows. All of them move 1 step at the same time. We asked the participants to determine at what step the human and a robot would collide (we make it easier by telling them there would be a collision). We presented two problems: problem 1 was optimized by GSC and had fewer vertices(14) but more branching vertices(3); problem 2 was optimized by BVC and had more vertices(16) and fewer branching vertices(1). Our objective was to see if it took less time for the same user to answer correctly for the problem with fewer branching vertices. Easier conflict detection should make movement in the space more seamless, as the human can better participate in the joint navigation in the space by avoiding conflicts.

We considered that there could be a wide disparity in the cognitive ability of people to handle or visualize combinatorial cases in paths, so we adopted the \emph{within-subject} experimental design to compare performance. The null hypothesis is that the mean time difference is 0 (no impact of branching vertices). If our assumptions were sound, then the mean time difference should be significantly less than 0 (takes less time with fewer branching vertices). We fix the human path, rather than let the participant choose their own path, as we wanted to control the study to determine the effect of branching vertices on the cognitive effort to predict a collision or not. 

Before we gave the problems, we first show an animated example of the human and robots moving in a different grid, with examples of a safe path and a path with a collision. Then we presented a practice problem with two robots and the same length for the human path. The subject had to get the practice problem correct before proceeding to the test problems; the practice problem was also on a different grid(problem setting). Then we presented the subjects with the two problems in a randomized order. The subject could not proceed to the next problem without getting the first problem correct; this was needed so we could compare the time taken, which in turn reflected relative complexity and cognitive effort. At the end of the experiment we debriefed participants by explaining the purpose of the experiment (to evaluate if one of the problems was easier to solve), and gave them the option to reach out to us for more information. 

To make the comparison fair, the human path in both problems was of the same length (7 steps), and overlapped with the robot positions the same number of times (3 positions) as can be seen in Figure \ref{fig:human_studies_problem}. In both problems, the human collides with a robot on step 5, which the participants ofcourse did not know. So, they had to rollout the path for the robots for the same number of steps before they saw a collision. We matched the human path and robot navigation-graph as closely as possible; this was possible since they came from the same initial graph. We made sure that atleast one branching-vertex in both problems were involved in determining when the human collided with the robot.
Thus the major difference is the effect of the additional branching vertices in problem 1, which would naturally create more paths to consider.

To minimize the carryover effect in within-subject experiments, we presented an animated example and a practice example to sufficiently prepare the user, and give less of a learning benefit when doing either of the test problems first. We also randomized the order of problems to reduce the impact of order on the data. All participants were from the Amazon Mechanical-Turk service \cite{Mturk_crowston2012amazon} and only filtered by ``masters" qualification for users (users that had a good track record).There were 36 subjects in total. We tracked the time taken and guesses made for each problem. Of the 36 data points, 7 were removed. Of the 7, 2 were removed as it was apparent they were randomly guessing (we recorded guesses and time of the guess). 4 were removed because they were extreme outliers, as in the subjects took in excess of 201 seconds for a problem. We computed the outlier cutoff using the interquartile rule \cite{IQR_outlier_rousseeuw2011robust}. By this rule, we add $1.5\times IQR$ (inter quartile range) to the $75{th}$ percentile value to get the upperbound; the lower bound similarly calculated from the $25{th}$ percentile value, and this was below zero and so ignored. Note that the outlier data points also supported our hypothesis as time taken for problem 2 was much less than problem 1, but we eliminated them due to them being extreme outliers which would hurt the paired t-test evaluation. The last remaining data point was removed because the user solved both problems in 2 seconds which implied they either knew the answers or repeated the test under a new ID. Once we have the time taken for solving each problem, we take the difference between the time taken for problem 1 and 2, and run a paired t-test \cite{paired_ttest_hsu2014paired}, we were looking for evidence at the 5\% significance level (p-value 0.05). The mean difference in time taken was -47.53 seconds, i.e. problem 2 on average takes less time. The statistical significance of this result comes from the paired t-test analysis done with the Scipy library in python \cite{scipy}. The p-value computed is 0.0006, which means we can reject the null hypothesis of no difference in time taken with high confidence. If we only consider the data with only problem 1 appearing first, the mean difference is -49.08 and the p-value is 0.0118. If we only consider the data with problem 2 appearing first, the mean difference is -46.5 and the p-value is 0.0185. So we can reject the null hypothesis, and we have strong confidence in the mean difference being lower in favor of fewer branching vertices. The average time to do problem 1 and 2 was 87 seconds and 38 seconds respectively. We certainly do not expect people to invest such amounts of time navigating around robots. The time taken was because we asked people to count the steps of the human and robot and indicate the earliest possible step at which they would collide. Counting to get the exact and earliest collision step, and having to consider combinations of path segments is why the time cost is higher. In reality, humans do not count and get the exact time or step of a path-conflict, only that we intuit there might be one and adjust our path. Our human-subjects problem was to test if the cognitive cost would be higher for predictions when there are more branching vertices, and the data supports this. We expect this relative cognitive cost to translate to humans navigating in the real world with robots.

\section{CONCLUSION}
In this paper we presented the problem of computing the navigation-graph for position-based predictability which can be used to determine how an AGV's grid is laid out. We defined measures for position-based predictability with justifications for them. Finally, we presented and evaluated a hill-climbing algorithm to minimize graphs for position-based predictability.
We conducted human-subject studies to verify that branching vertices do indeed make the problem of predicting the motion of the robots from their current position alone harder; we evaluated how quickly they could determine when two paths would intersect in two problems, one with fewer branching vertices (which came from our algorithm). We found the time taken to be significantly lower in favor of the navigation-graph with fewer branching vertices.

\section*{ACKNOWLEDGEMENTS}
This research is supported in part by ONR grants N00014- 16-1-2892, N00014-18-1- 2442, N00014-18-1-2840, N00014-9-1-2119, AFOSR grant FA9550-18-1-0067, DARPA SAIL-ON grant W911NF19- 2-0006 and a JP Morgan AI Faculty Research grant.



\bibliography{references}


\clearpage
\newpage


\begin{appendices}

\appendixpage
\addappheadtotoc

\section{Hill Climbing Pseudocode}

The pseudocode for our hill climbing approach is captured in Algorithms \ref{alg:graph_min_pred} and \ref{alg:replace_path} in the following pages. 
\begin{algorithm*}
  \caption{Graph Minimization For Online Predictability By Hill Climbing Search}
    \begin{algorithmic}[1]
        \Function{Graph\_Min}{$G=(V,E),tasks,cutoffMap,weightMap,restartCount,maxPopulation,CostFunc$}
      \State //NOTE: "Map" in the variable names refers to the data structure map, as in key->value mapping, not the navigation graph
        \State //In terms of the problem definition $tasks = T,cutoffMap = C, weightMap = W$
            \State $taskPathsMap$ = BUILD\_CANDIDATES($G, tasks, cutoffMap, maxPopulation$)
            \State $minimized\_graph$ = 
            \State \space \space \space HC\_SEARCH($G, tasks, taskPathsMap, cutoffMap, weightMap, restartCount,CostFunc$)\\
            \Return $minimized\_graph$
        \EndFunction
        \\\hrulefill
      \Function{Build\_Candidates}{$G=(V,E),tasks,cutoffMap,maxPopulation$}
        \State $taskPathsMap = \{\}$ //maps each $(i,g)\in T$ to a set of paths
        \For{$(i,g) \in tasks$}
            \State $copyG$ = COPY($G$) //make an copy of the graph to edit
            \For{$i=0$ to $maxPopulation$}
                \State $shortestPath$ = GET\_SHORTEST\_PATH($copyG,i,g$)
                 //then check if true cost in G is less than cutoff                 \If{LENGTH($G,shortestPath$)$<cutoffMap[(i,g)]$} 
                    \State $taskPathsMap[(i,g)] = taskPathsMap[(i,g)]\cup {shortestPath}$ //add to the paths set for this task
                    \For{$(v_1,v_2) \in shortestPath$}
                        \State $copyG[v_1][v_2] *= 2$ //double the path edges' weights to help get different paths in each iteration
                    \EndFor
                \EndIf
            \EndFor
        \EndFor
        \State \Return $taskPathsMap$
      
      \EndFunction
    \\\hrulefill
    \Function{HC\_SEARCH}{$G, tasks, taskPathsMap, cutoffMap, weightMap, restartCount,CostFunc$}
    \State $bestMinCost = \infty$ // set to the largest 32 or 64 bit number
    \State $bestPathsMap = \{\}$ //track the best paths for each task $t = (i,g)$
    \For{$i = 0$ to $restartCount$} // this loop is for random restarts to improve search
        \State $chosenPathsMap = f:{t \in tasks} \mapsto RANDOM\_CHOICE(taskPathsMap[t])$ 
        \State $roundBestMinCost = $ CostFunc($chosenPathsMap,weightMap$)
        \While{True}
            \State $prevCost = roundBestMinCost$
            \For{$t \in tasks$}
                \State $newCost,newPathMap$ = REPLACE\_PATH(\\
                \hspace{35mm}   $t,G, tasks, currPathsMap, taskPathsMap, cutoffMap, weightMap,CostFunc$)
                \If{$newCost < roundBestMinCost$}
                    \State $roundBestMinCost = newCost$
                    \State $chosenPathsMap = newPathMap$
                \EndIf                                                      
            \EndFor 
            \If{$prevCost == roundBestMinCost$}
                \State \emph{break} //Terminate current round when no improvement is seen
            \EndIf
        \EndWhile
        \If{$roundBestMinCost < bestMinCost$}
            \State $bestMinCost =roundBestMinCost$
            \State $bestPathsMap = currPathsMap$
        \EndIf
    \EndFor

    \State $minimizedGraph$ = BUILD\_GRAPH\_FROM\_PATHS($bestPathsMap$) \State //return a graph with all the vertices and edges from the given paths
    \State \Return $minimizedGraph$
    \EndFunction
\end{algorithmic}
\label{alg:graph_min_pred}
\end{algorithm*}
\restylefloat{algorithm}
\begin{algorithm*}
  \caption{Replacing paths to reduce the number of branching vertices}
    \begin{algorithmic}[1]
      \Function{REPLACE\_PATH}{ $t,G, tasks,chosenPathsMap,taskPathsMap,cutoffMap, weightMap,CostFunc$}
        \State $currBestPath = chosenPathsMap[t]$
        \State $minCost = $ CostFunc($chosenPathsMap,weightMap$) 
        \State $chosenPathCost = $ ADD\_EDGE\_WEIGHTS($currBestPath,G$)//sum with edge weights in graph G
        \State $editablePathMap = $ COPY($chosenPathsMap$) //create a copy to hold potential replacement paths
        \For{$altPath$ in $taskPathsMap[t]$}
            \State $editablePathMap[t] = altPath$//replace single path
            \State $newCost = $ CostFunc($editablePathMap,weightMap$)
            \State $newPathCost = $ ADD\_EDGE\_WEIGHTS($altPath$)
            \If{$newCost < minCost$}
                \State $minCost = newCost$
                \State $chosenPathCost = newPathCost$
                \State $currBestPath = altPath$
            \EndIf
            \If{$newCost == minCost \And newPathCost < chosenPathCost$}
                \State $minCost = newCost$ //the graph cost is the same, but the indiv path is shorter, then replace
                \State $chosenPathCost = newPathCost$
                \State $currBestPath = altPath$
            \EndIf
            \State $editablePathMap = $ COPY($chosenPathsMap$) //reset this data structure for next iteration
        \EndFor
        \State $editablePathMap[t] = currBestPath$  //replace path for task "t" with the best path found
        \State \Return ($minCost,editablePathMap$)
      
      \EndFunction
\end{algorithmic}
\label{alg:replace_path}
\end{algorithm*}




\section{Additional Examples of Navigation-Graph Computation }

\begin{figure*}[ht]
\centering
     \includegraphics[width=1.0\linewidth]{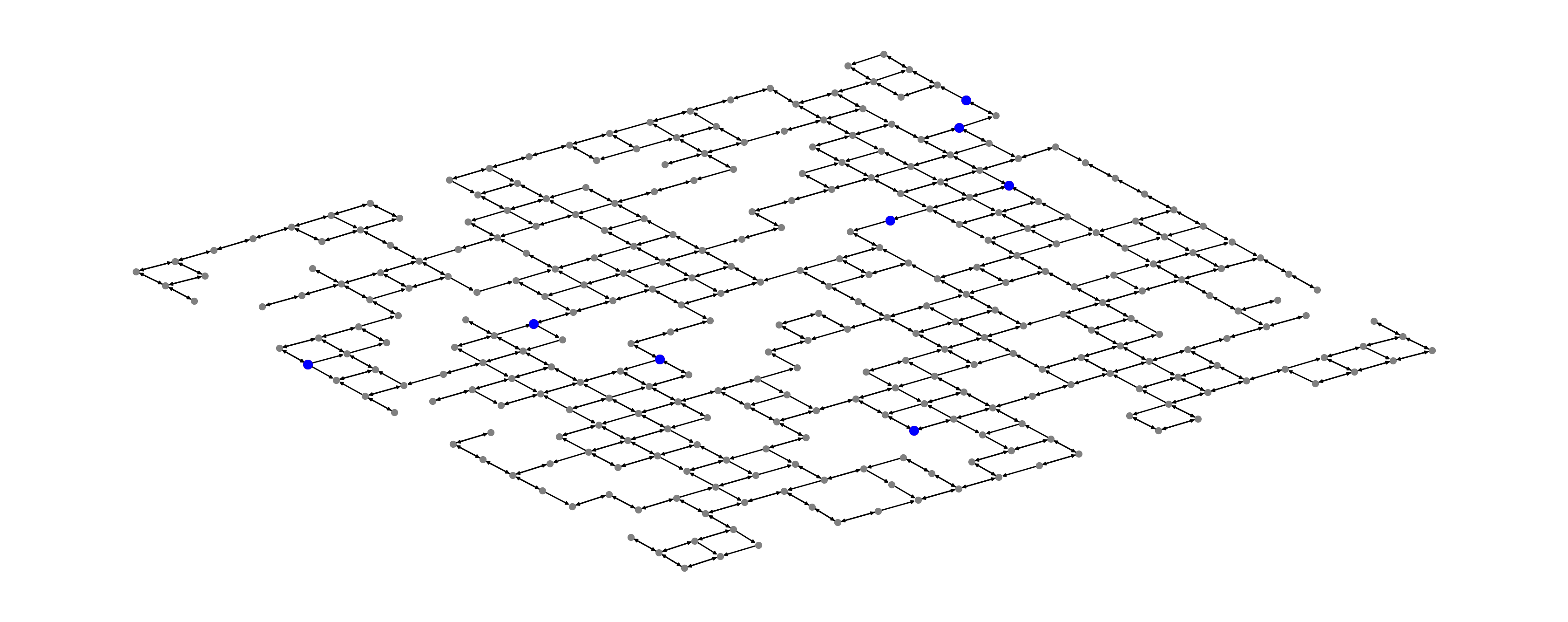}
      \caption{Example 1 of a 20x20 grid with 20\%of nodes and 20\% of edges randomly dropped, and 8 terminal vertices}
      \label{fig:20_20_1}
\end{figure*}

\begin{figure*}[ht]
\centering
     \includegraphics[width=1.0\linewidth]{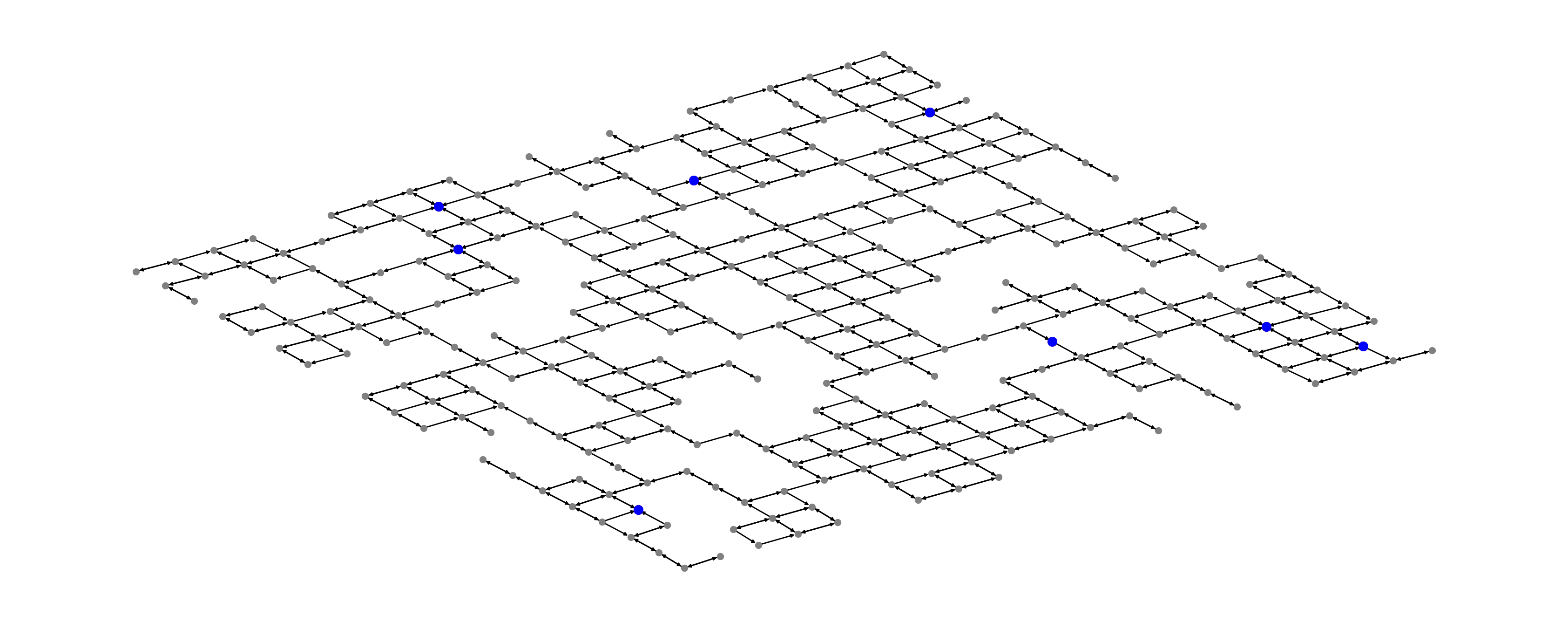}
      \caption{Example 2 of a 20x20 grid with 20\%of nodes and 20\% of edges randomly dropped, and 8 terminal vertices}
      \label{fig:20_20_2}
\end{figure*}

\begin{figure*}[ht]
\begin{subfigure}{.5\textwidth}
  \centering
  \includegraphics[width=1.0\linewidth]{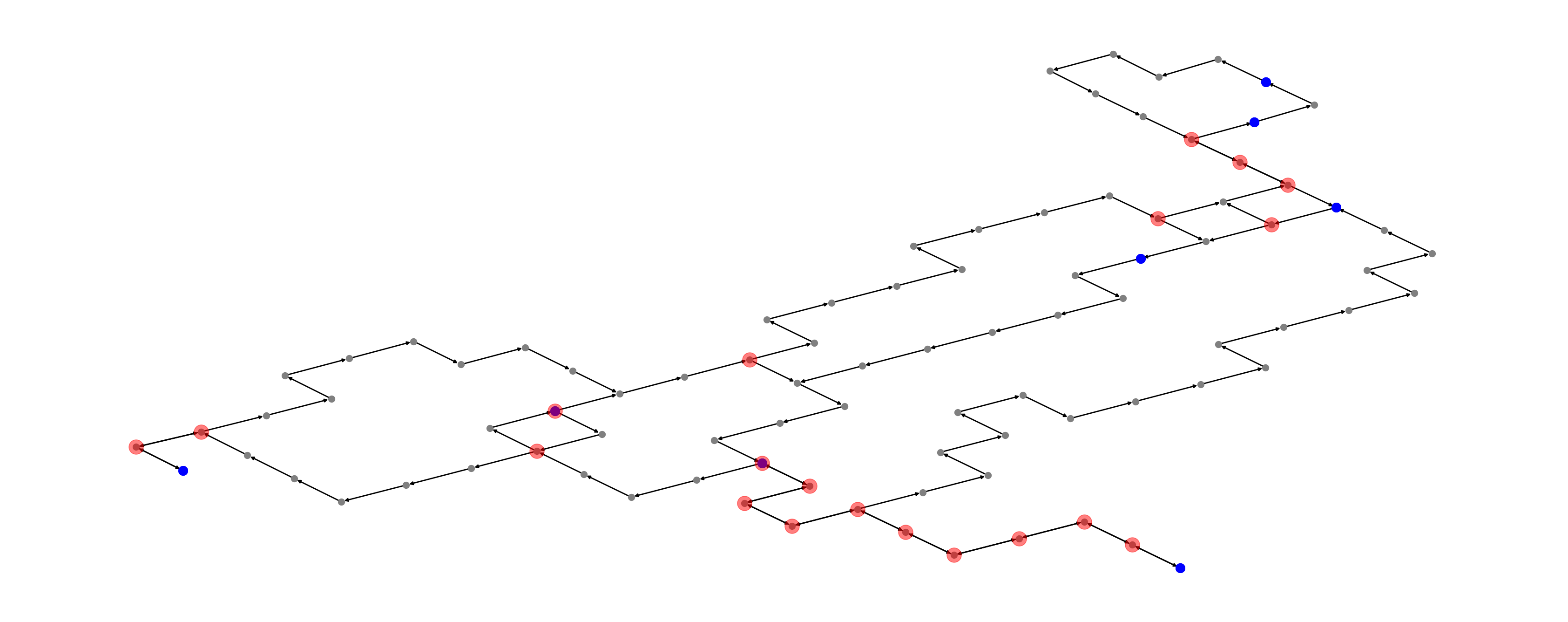}
  \caption{Navigation Grid Computed With GSC cost function \\ for Example 1 Graph in Figure \ref{fig:20_20_1}}
  \label{fig:20_20_1_GSC}
\end{subfigure}%
\begin{subfigure}{.5\textwidth}
  \centering
  \includegraphics[width=1.0\linewidth,height=3cm]{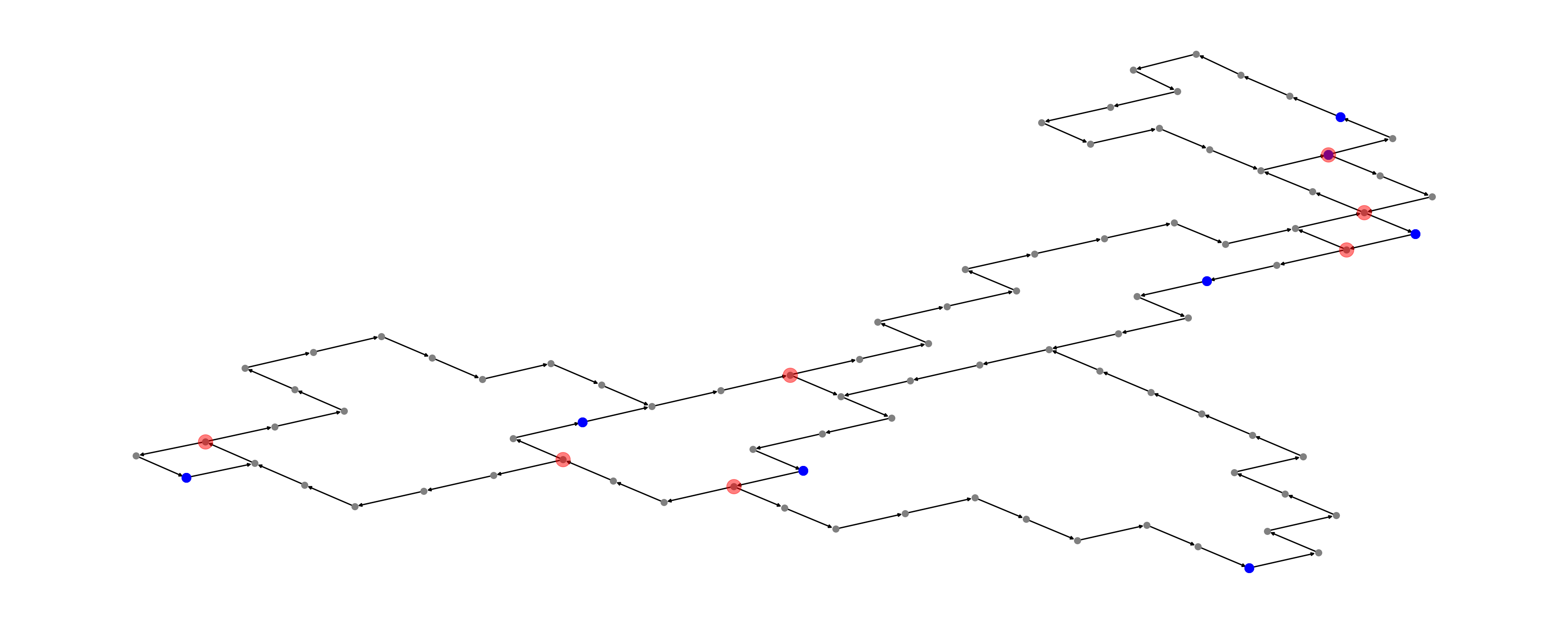}
  \caption{Navigation Grid Computed With BVC cost function \\ for Example 1 Graph in Figure \ref{fig:20_20_1} with significantly fewer branching vertices (red) than in Figure \ref{fig:20_20_1_GSC}}
  \label{fig:20_20_1_BVC}
\end{subfigure}%
\caption{Navigation graphs computed on the full graph in Figure \ref{fig:20_20_1}}
\end{figure*}

\begin{figure*}[ht]
\begin{subfigure}{.5\textwidth}
  \centering
  \includegraphics[width=1.0\linewidth]{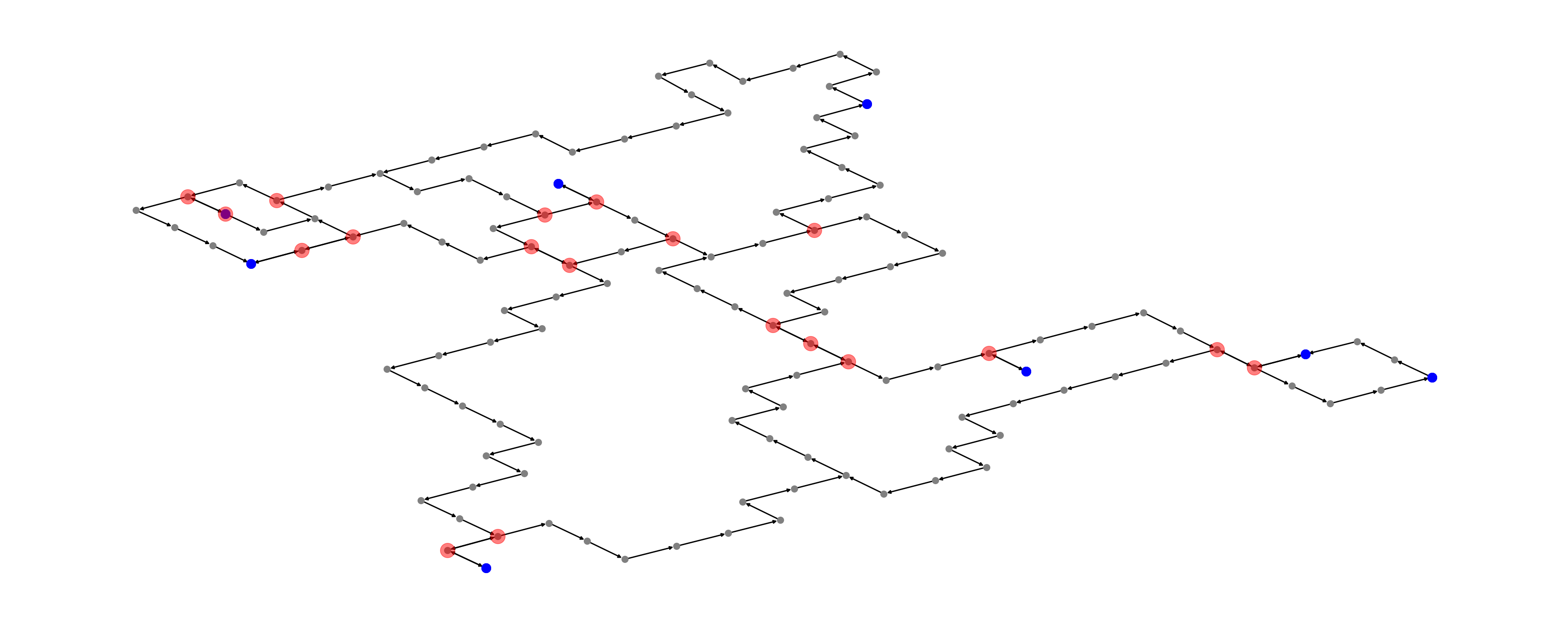}
  \caption{Navigation Grid Computed With GSC cost function \\ for Example 2 Graph in Figure \ref{fig:20_20_2}}
  \label{fig:20_20_2_GSC}
\end{subfigure}%
\begin{subfigure}{.5\textwidth}
  \centering
  \includegraphics[width=1.0\linewidth,height=3cm]{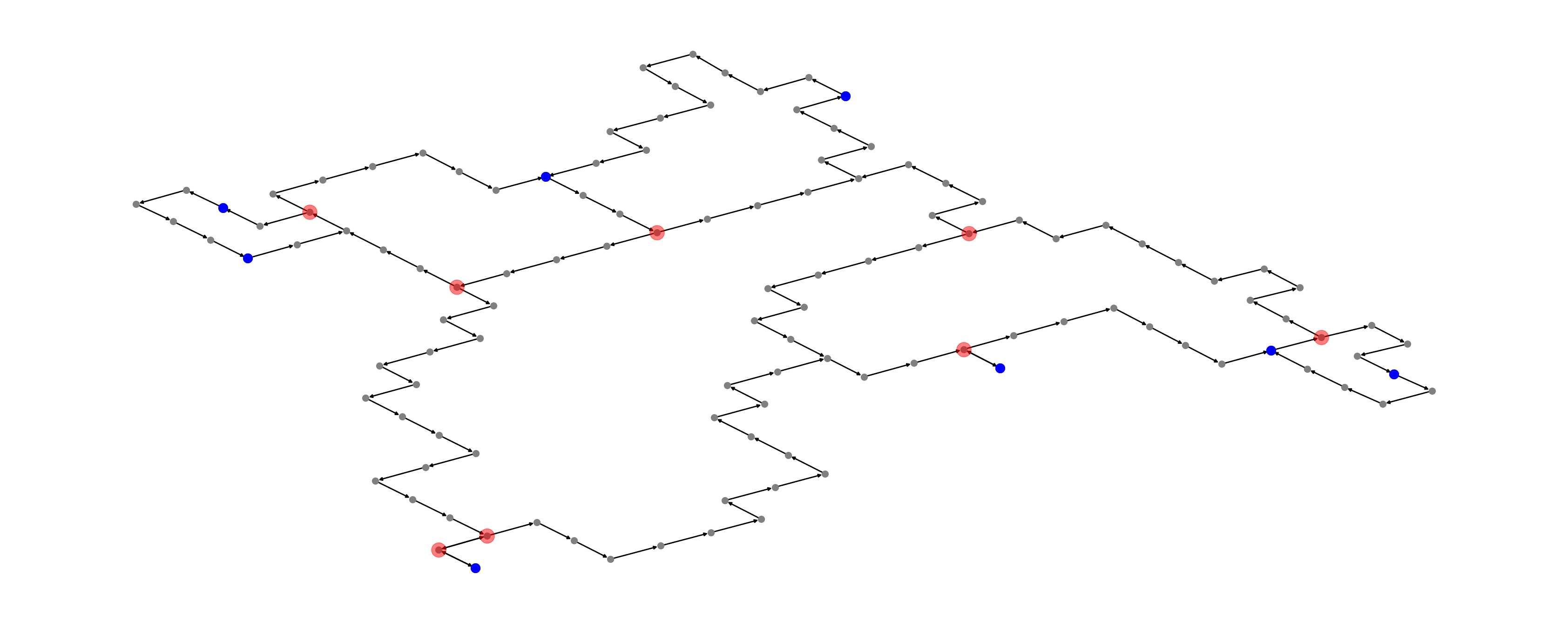}
  \caption{Navigation Grid Computed With BVC cost function \\ for Example 2 Graph in Figure \ref{fig:20_20_2} with significantly fewer branching vertices (red) than in Figure \ref{fig:20_20_2_GSC}}
  \label{fig:20_20_2_BVC}
\end{subfigure}%
\caption{Navigation graphs computed on the full graph in Figure \ref{fig:20_20_1}}
\end{figure*}

We provide two additional examples of graphs and the associated navigation-graphs computed on them using GSC and BVC cost functions. Red vertices are branching vertices, and blue vertices are terminal vertices. These two examples are in Figures \ref{fig:20_20_1},\ref{fig:20_20_1_GSC},\ref{fig:20_20_1_BVC} and \ref{fig:20_20_2},\ref{fig:20_20_2_GSC},\ref{fig:20_20_2_BVC} respectively

\end{appendices}

\end{document}